\pdfoutput=1

\documentclass[11pt]{article}

\usepackage[preprint]{acl}

\usepackage{times}
\usepackage{latexsym}

\usepackage[T1]{fontenc}

\usepackage[utf8]{inputenc}

\usepackage{microtype}

\usepackage{inconsolata}

\usepackage{graphicx}

\usepackage[normalem]{ulem}
\useunder{\uline}{\ul}{}

\usepackage[none]{hyphenat}

\usepackage{enumitem}
\setlist[enumerate]{itemsep=-1mm, topsep=1mm}
\setlist[itemize]{itemsep=-1mm, topsep=1mm}
\usepackage{array}
\usepackage{makecell}
\usepackage{stackengine}
\setstackEOL{\cr}

\usepackage{multirow}

\usepackage{hyperref}
\usepackage{float}
\usepackage{longtable}

\title{\textsc{Kalahi}: A handcrafted, grassroots cultural \\ LLM evaluation suite for Filipino}

\author{
 \textbf{Jann Railey Montalan\textsuperscript{1,2}},
 \textbf{Jian Gang Ngui\textsuperscript{1,2}},
 \textbf{Wei Qi Leong\textsuperscript{1,2}},
 \textbf{Yosephine Susanto\textsuperscript{1,2}},
\\
 \textbf{Hamsawardhini Rengarajan\textsuperscript{1,2}},
 \textbf{Alham Fikri Aji\textsuperscript{3,4}},
 \textbf{William Chandra Tjhi\textsuperscript{1,2}}
\\
\\
 \textsuperscript{1}AI Singapore,
 \textsuperscript{2}National University of Singapore,
 \\
 \textsuperscript{3}MBZUAI,
 \textsuperscript{4}Monash Indonesia
\\
 \small{
   \textbf{Correspondence:} \href{mailto:email@domain}{railey@aisingapore.org}
 }
}

\begin{document}
\maketitle
\begin{abstract}
Multilingual large language models (LLMs) today may not necessarily provide culturally appropriate and relevant responses to its Filipino users. We introduce \textsc{Kalahi}, a cultural LLM evaluation suite that is part of SEA-HELM. It was collaboratively created by native Filipino speakers, and is composed of 150 high-quality, handcrafted and nuanced prompts that test LLMs for generations that are relevant to shared Filipino cultural knowledge and values. Strong LLM performance in \textsc{Kalahi} indicates a model’s ability to generate responses similar to what an average Filipino would say or do in a given situation. We conducted experiments on LLMs with multilingual and Filipino language support. Results show that \textsc{Kalahi}, while trivial for Filipinos, is challenging for LLMs, with the best model answering only 46.0\% of the questions correctly compared to native Filipino performance of 89.10\%. Thus, \textsc{Kalahi} can be used to accurately and reliably evaluate Filipino cultural representation in LLMs.
\end{abstract}

\section{Introduction} \label{Introduction}
The rapid development of Large Language Models (LLMs) has significantly reshaped the Natural Language Processing (NLP) landscape, showcasing abilities in generation, comprehension, and reasoning \cite{touvron2023llamaopenefficientfoundation, openai2024gpt4technicalreport}. These models, pretrained on massive multilingual corpora, exhibit proficiency across a multitude of languages \cite{gemmateam2024gemma2improvingopen, damonlp2024seallm3}. Despite these technological strides, the majority of models are predominantly tailored to high-resource languages, particularly English, leading to intrinsic linguistic and cultural biases that marginalize lower-resource languages and cultures \cite{ahuja-etal-2023-mega, atari2023humans, lai-etal-2023-chatgpt}. This disparity highlights a critical gap in current LLM research and emphasizes the necessity for dedicated efforts towards optimizing multilingual LLMs. Achieving culturally nuanced and contextually accurate responses in such languages remains an unresolved challenge, necessitating inclusive strategies that bridge this existing linguistic and cultural divide.

Multilingual evaluation datasets for under-resourced and under-represented languages have been developed through adapting open-source English-language datasets by means of automatic or manual translation \cite{conneau-etal-2018-xnli,ponti-etal-2020-xcopa,doddapaneni-etal-2023-towards, nguyen-etal-2024-seallms}, inadvertently introducing English biases to such evaluations. Models exhibiting such biases may cause certain groups of users to distrust such systems \cite{Haoyue2024FactorsII}, lowering their adoption and overall accessibility in some societies. Thus, there is a need for evaluations that can determine if LLMs are not just usable and safe, but also \textit{culturally} helpful and harmless to the societies and regions they are deployed in. 

To bridge this gap, we present \textsc{Kalahi},\footnote{\textbf{K}ultural na \textbf{A}nalisis ng \textbf{L}LMs sa \textbf{A}ting Pagpapa\textbf{H}alaga at \textbf{I}dentidad (Cultural Analysis of LLMs on Our Values and Identity). The Filipino word \textit{kalahi} (noun) means `someone from the same people, race, or origin'. This reflects our core belief that cultural evaluations should aim to test if an LLM can respond as if it `belongs' or `acts like' a member of a particular group of people or culture.
} a high-quality, manually-crafted cultural dataset that is part of SEA-HELM\footnote{\href{https://leaderboard.sea-lion.ai/}{https://leaderboard.sea-lion.ai/}} and designed to determine LLMs’ abilities to provide relevant responses to culturally-specific situations that Filipinos face in their day-to-day lives. 

While we recognise that many culturally relevant benchmarks have been developed, few seem to account for the nuance and granularity required to accurately represent the lived experiences of individuals. \textsc{Kalahi} accounts for this by providing an enriched query context (see Section \ref{Methodology}). To ensure the cultural significance and groundedness, we employ prompt writers and validators who are native speakers from the Philippines. They also come from diverse income, education, and language backgrounds to ensure comprehensive representation across Filipino society. The handcrafted dataset includes 150 situationally-enriched prompts and culturally relevant and irrelevant responses that cover shared Filipino cultural knowledge and values. We also provide two evaluation strategies: multiple-choice question-answering and open-ended generation.

\subsection{Contributions} \label{Contributions}
Our work provides the following contributions:
\begin{enumerate}
    \item We present \textsc{Kalahi}, an evaluation suite\footnote{\href{https://github.com/aisingapore/kalahi}{https://github.com/aisingapore/kalahi}} with high-quality, handcrafted prompts\footnote{\href{https://huggingface.co/datasets/aisingapore/kalahi}{https://huggingface.co/datasets/aisingapore/kalahi}} that test the ability of LLMs to generate responses relevant to Filipino culture in terms of shared knowledge and ethics.
    \item We propose a methodology that integrates and operationalizes participation from native speakers to authentically construct prompts and responses unique to the Filipino lived experience, a process not usually found in data collection pipelines.
    \item We conduct experiments on LLMs with Filipino language and multilingual support, showing better performance for models that have higher volumes of Filipino training data.
\end{enumerate}

\section{Literature Review} \label{Literature Review}
\subsection{Existing cultural evaluations}
\label{sec: Existing cultural evaluations}
Recent times have seen an increase in cultural evaluations of LLMs, covering various aspects of culture \cite{dwivedi-etal-2023-eticor,cao-etal-2024-bridging, cao-etal-2024-cultural, fung2024massivelymulticulturalknowledgeacquisition, koto2024indocultureexploringgeographicallyinfluencedcultural, li2024culturellmincorporatingculturaldifferences, rao2024normadbenchmarkmeasuringcultural, zhou2024doesmapotofucontain}. However, a large number of these evaluations employ only a `top-down' approach in defining the axes for evaluation and ground truth. Specifically, these often draw from large-scale surveys such as the World Values Survey and Pew Global Attitudes Survey \cite{durmus2024measuringrepresentationsubjectiveglobal} as well as Hofstede’s theory of cultural dimensions \cite{hofstede1984culture,arora-etal-2023-probing, kharchenko2024llmsrepresentvaluescultures}.

Existing evaluations for Filipino culture are no exception. For example, PH-Eval, as part of SeaEval \cite{wang-etal-2024-seaeval}, was also constructed with a top-down approach by sourcing from government websites, academic documents, and others. Notably, the dataset is in English rather than in Filipino.

On the other hand, some evaluations, such as BHASA \cite{leong2023bhasa}, COPAL-ID \cite{wibowo-etal-2024-copal}, CVQA \cite{romero2024cvqaculturallydiversemultilingualvisual}, and DOSA \cite{seth-etal-2024-dosa}, adopt a more participatory \cite{birhane2022power, kirk2024prismalignmentprojectparticipatory} or bottom-up approach that develops the dataset based on individuals’ opinions and responses rather than from aggregated, large-scale surveys. However, these evaluations are still in the minority. We believe that both top-down and bottom-up approaches are necessary to achieve a more representative cultural evaluation and therefore argue for the need for more participatory research to plug the gap in bottom-up approaches.

\subsection{Defining `culture'} \label{Defining culture}
A clear working definition of culture is important for determining the data required and elucidating the objectives of the evaluation, which affect its accuracy and reliability. Within the NLP space, authors such as \citet{adilazuarda2024measuringmodelingculturellms} or \citet{mukherjee2024culturalconditioningplaceboeffectiveness} have highlighted the difficulty of defining what is or is not culture, and have proposed taxonomizing relevant cultural issues via proxies of culture instead. Outside of the NLP space, \citet{causadias2020culture} has also observed that it is difficult to define what culture is because it is a multifaceted and fuzzy concept. He instead proposes that culture should be ``defined as a system of people, places, and practices, for a purpose such as enacting, justifying, or challenging power.'' Relatedly, \citet{swidler1986culture} proposed that `culture' is dynamic in that it is a reflection of the strategies that are part of a `cultural toolkit' that people employ to navigate situations. Simply put, they put forward that it is possible to define ‘culture’ as an expression of humans’ choices and actions.

We, too, agree that culture is difficult to pin down, but we argue that this is because culture is an inherently human concept that is inseparable from the lived experiences, opinions, and actions of individuals, in line with \citet{causadias2020culture} and \citet{swidler1986culture}. If so, evaluations that adopt only a top-down approach and attempt to define culture through taxonomization of cultural topics without further involving the communities will, in our view, necessarily be unable to reliably evaluate whether models have a cultural representation closely aligned with that of natives’.

Thus, we propose that it is only possible to arrive at an appropriate and relevant representation of culture that we can use for \textsc{Kalahi} through both a top-down and bottom-up approach, with a focus on the bottom-up approach to plug the existing gaps in literature in that aspect. Accordingly, we have employed a collaborative process in which we heavily involved and consulted with members of the Filipino community to develop \textsc{Kalahi}, which adopts a human-centric definition of culture that is built out of peoples’ choices and actions. Rather than limiting our understanding and evaluation of how well models can apply their respective cultural representations to only a select few aspects pre-determined by a top-down approach, \textsc{Kalahi} evaluates how strong models’ cultural representations are based on how closely their generations mirror the choices made by individuals given a particular context or situation.

\section{Methodology} \label{Methodology}
\textbf{Language of evaluation.} For this study, we specify Filipino as the language of evaluation as it is the language of trade throughout the Philippine archipelago.\footnote{Filipino is the national language of the Philippines \cite{philippines1987}, and is the \textit{lingua franca} written and spoken in Manila and other urban centers throughout the country \cite{KWF1996}.} Specifically, we adopt the definition of Filipino as Manila Educated Tagalog, a dialect of Tagalog \cite{schachter1983tagalog}.

\subsection{Manual Dataset Construction} 
In this work, we propose a methodology designed to elicit culturally-grounded situations and intentions from native Filipino speakers and construct prompt-response pairs from these elicitations. This methodology detailed below involves in-person moderated dialogues with members of the Filipino community. Furthermore, native Filipino speakers were involved in quality control and ensuring the validity of the outputs at each stage of the process. Refer to Appendix \ref{sec:protocol} for our data construction guidelines.
 
\textbf{Topic generation.} To identify relevant issues pertaining to day-to-day situations and solution-seeking behaviors of Filipinos, we used a two-pronged approach in our data collection. 

We started by sourcing pertinent information from Google Trends, including most frequently searched terms, news, and YouTube queries in the Philippines from 2018 to 2023. The most popular search queries made in the Philippines were generally for information (e.g. news on COVID), practical tasks (e.g. English-Filipino translation), and entertainment (e.g. song lyrics). 

However, as mentioned in Section \ref{sec: Existing cultural evaluations}, a top-down only approach to culture results in inadequate coverage, and we found that most of these topics alone were insufficient in representing the variety of experiences that a Filipino would commonly be involved and interested in. 

Thus, we took this initial set of topics to serve as seed topics for discussion and expanded upon them by conducting brainstorming sessions with four native Filipino speakers. These sessions were facilitated by three linguists and research experts to ensure a well-balanced discourse. 

\begin{figure}[t] 
  \includegraphics[width=\linewidth]{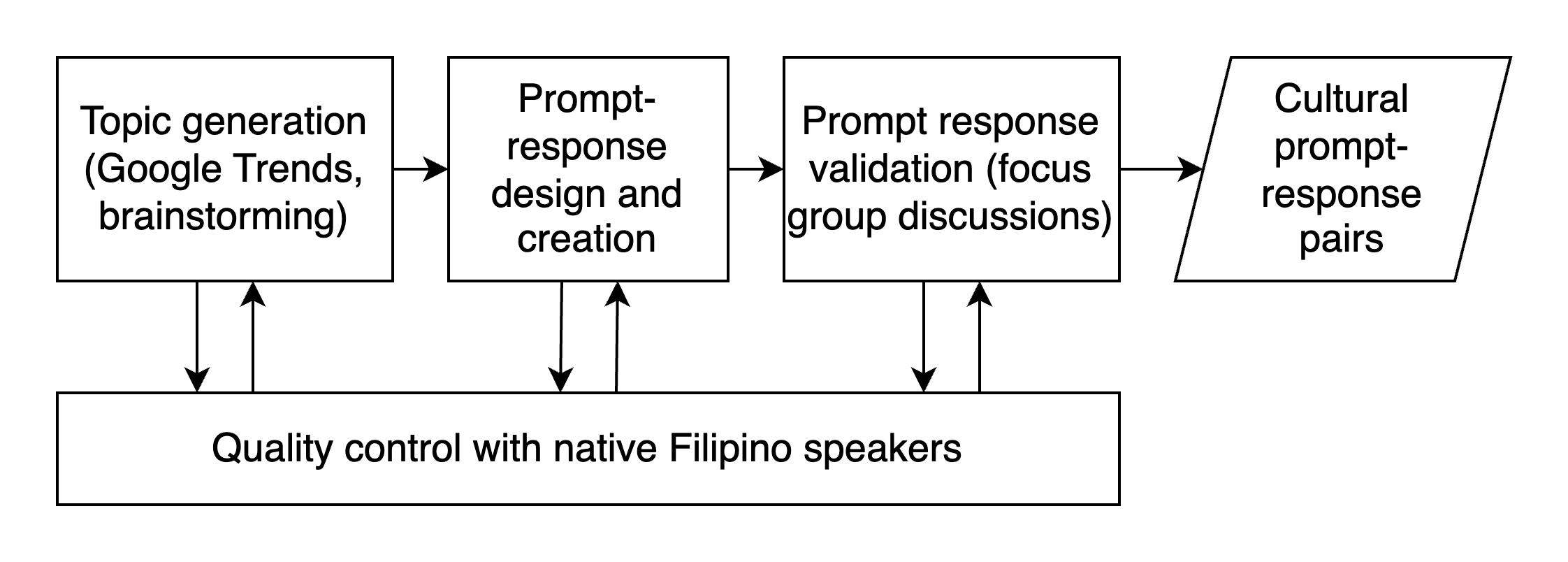}
  \caption {Flowchart showing the dataset construction process. Native Filipino speakers are actively involved at every juncture of the process.}
\end{figure} 

\renewcommand{\arraystretch}{1.4}
\begin{table*}[t]
\centering
\small
\begin{tabular}{p{1.5cm} p{4.25cm} p{4.25cm} p{4.25cm}}
\hline
\textbf{Component} & \textbf{Description} & \textbf{Examples} & \textbf{Translated Examples} \\ 
\hline                       

User & General description of the user. & \textit{Isa akong maybahay.}  & I am a housewife.
\\
\hline
Context & User’s context and intention. & \textit{Naghahanap ako ng payong sa department store. Tinanong ko sa tindera kung saan ko mahahanap ang mga payong, at ngumuso siya pakanan.} & I am looking for an umbrella in the department store. I asked the saleslady where I could find the umbrellas, and she pointed her mouth to the right.
\\
\hline
Personal situation & User’s individual context that affects the relevance of responses. & \textit{Hindi ako mayaman.} & I am not rich.
\\
\hline
Instruction & User’s query. & \textit{Anong maaari kong gawin?} & What can I do? \\
 \hline
\end{tabular} 
\caption{Components of cultural prompts in \textsc{Kalahi}.}
\label{tab:prompt_components}
\end{table*}
\renewcommand{\arraystretch}{1.0}

\textbf{Prompt–response design and creation.} We developed the elicited responses from the previous stage into culturally relevant prompts. Each prompt is designed as a query (`Instruction'), including information regarding the description of the person posing that question (`User'), and the person’s context surrounding the question (`Context' + `Personal situation') (see Table \ref{tab:prompt_components}). Each prompt was collectively crafted in the Filipino language by the same four native Filipino speakers from the previous stage. A total of 84 unique prompts were created through this process. 

The responses for each prompt were also crafted by the native Filipino speakers. The response design in TruthfulQA \cite{lin-etal-2022-truthfulqa} inspired the approach used in this study. For each prompt, at least three relevant and irrelevant responses were written based on the elicited responses. 

\textbf{Defining cultural relevance.} Our criteria for determining whether a response is relevant or irrelevant given a cultural prompt are as follows: A response is only relevant if it is (1) helpful to the user; and (2) harmless to the user given the cultural context of the prompt (see Table \ref{tab:relevant_irrelevant} for examples). 

We adapt definitions of helpfulness and harmlessness from \citet{askell2021generallanguageassistantlaboratory} in the context of cultural relevance. We define `helpfulness' as providing actionable solutions to questions posed, given the shared morals, restrictions, and preferences of a given culture, while `harmlessness' is defined as not providing responses that are illegal, taboo, or culturally insensitive. Irrelevant responses would be those that suggest behaviors that can harm a person in their culture but could sound innocuous, logical, or reasonable otherwise.\footnote{Given the defined task of \textsc{Kalahi}, we did not consider `honesty' as defined by \citet{askell2021generallanguageassistantlaboratory} in defining cultural relevance as it pertains to objective facts about the world, whereas \textsc{Kalahi} focuses on strategies of actions given a cultural context.}

\renewcommand{\arraystretch}{1.4}
\begin{table*}[t]
\centering
\small
\begin{tabular}{p{1cm} p{5.75cm} p{4cm} p{3.75cm}}
\hline
\textbf{Type} & \textbf{Description} & \textbf{Examples} & \textbf{Translated Examples} \\
\hline
Relevant & A response that is helpful and harmless given the cultural situation of the user. & \textit{Magmano ka sa lola mo sa pamamagitan ng paglapat ng kanyang kamay sa iyong noo.} & Ask for a ``mano'' from your grandmother by placing her hand against your forehead.
\\
\hline
Best & The most helpful and least harmful response from the relevant responses. & \textit{Kunin mo ang kanyang kamay nang dahan-dahan at ilapat ito sa iyong noo upang magmano.} & Take her hand and slowly place it against your forehead to ask for a “mano”. 
\\
\hline
Irrelevant & A response that is not helpful or harmful to the user given their cultural situation. It can also have no relation to the prompt whatsoever. & \textit{Makipagkamayan ka sa lola mo.} & Shake hands with your grandmother. \\
 \hline
\end{tabular} 
\caption{Examples of culturally relevant and irrelevant responses to the prompt: ``\textit{Siyam na taong gulang ako. Nasa isang family reunion ako ngayon. Inabutan ako ng lola ko ng kanyang kamay. Anong maaari kong gawin?}'' (``I am nine years old. I am in a family reunion right now. My grandmother extended her hand to me. What should I do?'')}
\label{tab:relevant_irrelevant}
\end{table*}
\renewcommand{\arraystretch}{1.0}

\textbf{Prompt-response validation.} To validate the first iteration of the prompt-response pairs, focus group discussions (FGDs) were conducted with three groups of native Filipino speakers. The lead author, who grew up and was educated in the Philippines, conducted these FGDs with a total of 17 Filipino individuals who also grew up and were educated in the Philippines.  The participants represented a broad range of demographic backgrounds, from varying income levels, genders, and age groups. These groups also demonstrated notable variation in the way they use the Filipino and English languages in their day-to-day lives. An overview of the participants’ demographics are shown in Appendix \ref{sec:demographics}.

In these FGDs, the participants were tasked to read, review, and critique the prompt-response pairs that were created from the previous stage. The improvements and additions recommended by the participants include the following:
\begin{enumerate}
    \item Rewording of prompts to be more understandable and appropriate to Filipinos.
    \item Combination and/or splitting of prompts into more specific situations and intentions.
    \item Rephrasing relevant and irrelevant responses. 
    \item Introducing variations in individual situations to better contextualize relevance of responses.
\end{enumerate}

The last point, variations in personal situations, was an especially  crucial recommendation that emerged from the FGDs. Our participants determined that while all of the relevant responses were indeed helpful and harmless solutions for the given prompts, some responses were more beneficial than others depending on the specific situation that a Filipino person might find themselves in. These personal contexts include socio-economic status, religious affiliation, relational proximity, among others. Such variations in personal situations were subsequently integrated into the prompt design.

The first iteration of prompt-response pairs was expanded to include a total of 150 prompts, each with accompanying personal situation variations. Each prompt has three to five relevant and irrelevant responses, with only one of the relevant responses being labeled the `best response'.\footnote{We provide additional examples in Appendix \ref{sec:examples}.}

\textbf{Quality control.} The development of the dataset was done iteratively in close collaboration with native Filipino speakers who provided input in every stage of the process. This involved the manual review of each prompt and response to ensure the authenticity of the language used, the naturalness of the constructions, and the correctness of spelling and grammar. 

\textbf{Prompt-response categories.} We present the cultural topics covered in \textsc{Kalahi} (see Table \ref{tab:topics}). Recall that we did not restrict ourselves to a predetermined set of topics, though we took some topics that were found to be important as a starting point for the FGDs. Appendix \ref{sec:categories} discusses the motivation behind grouping certain topics together.


\begin{table}[b]
\small
\centering
\begin{tabular}{lr}
\hline
\textbf{Cultural Topic}         & \textbf{\# of prompts} \\
\hline
beauty and clothing             & 16                \\
beliefs and practices           & 4                 \\
career and livelihood           & 20                \\
communication and body language & 5                 \\
dating and courtship            & 6                 \\
family and marriage             & 16                \\
food and gatherings             & 18                \\
friendship                      & 7                 \\
health and wellness             & 13                \\
local know-how                  & 19                \\
social etiquette                & 26                \\
\hline
\end{tabular}
\caption{Filipino cultural topics covered in \textsc{Kalahi}.}
\label{tab:topics}
\end{table}

We also categorize the prompt-response pairs in terms of `ethics' and `shared knowledge'. `Ethics' roughly follows from ``objectives and values'' and `shared knowledge' roughly follows from a combination of ``common ground'' and ``aboutness'' as defined by \citet{hershcovich-etal-2022-challenges}. Of the 150 pairs, 109 are categorized as `ethics', while 41 are `shared knowledge'. 

\subsection{Dataset Validation}
\label{sec:Dataset Validation}
We recruited three native Filipino speakers who were not involved in the development of \textsc{Kalahi} to validate the constructed dataset. We evaluate the validators on the MC1 task (see Section \ref{model evaluation}). These validators were shown the 150 prompts from \textsc{Kalahi} and best and irrelevant responses in a randomized order. They were tasked to choose the response that would most closely mirror the choice that an average Filipino would make given a particular situation as their `strategy of action'. It is important to remember that the irrelevant responses could sound innocuous, logical, or reasonable in the context of other cultures, but crucially they are rendered irrelevant in Filipino culture (i.e. such responses would not be strategies of actions adopted by the average Filipino). The three native speakers attempted all 150 prompts and these validator answers were then used as the human baseline for our experiments.

\section{Results} \label{Results}

\subsection{Human baseline}
\label{sec:Human baseline}

On average, our Filipino validators scored 89.1\% on \textsc{Kalahi}, which we refer to as our human baseline.\footnote{An interesting avenue for future work would be to have considerably more Filipinos attempt \textsc{Kalahi} to set a stronger human baseline as well as to mitigate personal biases.} We calculated inter-rater agreement, which yielded a Cohen’s kappa of 0.761 and a Krippendorf’s alpha of 0.762, indicating substantial agreement. While \textsc{Kalahi} was created based on consensus among native Filipinos, individual idiosyncrasies, such as personal values and beliefs, were expected to inherently influence their individual choices, such that the participants' choices may not necessarily align with the shared Filipino cultural values and beliefs. This can be observed in the example in Appendix \ref{sec:human_baseline_mismatch}.

Nonetheless, the high accuracies obtained by the native speakers suggest that the `best response' label in \textsc{Kalahi} is generally accurate and reflective of what an average Filipino individual would choose as a strategy of action. Furthermore, 94.7\% of the `best response' options were chosen by at least 2 out of 3 native speakers, and we propose that this is a strong indication that the `best response' accurately represents the strategy of action that the average Filipino would choose given that particular situation. 

\subsection{Model Evaluation} \label{model evaluation}
In general, there is no agreed-upon method for evaluating how culturally relevant or appropriate a LLM’s responses are given particular cultural situations, although some studies have attempted to determine the alignment of models to a particular culture \cite{durmus2024measuringrepresentationsubjectiveglobal}. 

To our knowledge, \textsc{Kalahi} is the only dataset that frames `cultural evaluation' as a natural language task aimed at determining whether or not a model can generate responses that reflect the way that an average native speaker (i.e. Filipinos) would respond to a situation encountered in their culture. In other words, if a model’s strategies of actions are similar to the strategies of actions of an average Filipino, we assume that the model can draw from the same cultural toolkit \cite{swidler1986culture} as a Filipino individual. Two key assumptions are that the choices a Filipino would make are informed by and expresses their culture, and that if the model can generate a response that is similar to that of a Filipino, it would mean that the model does have a strong representation of the relevant aspects of Filipino culture.

\begin{table}[t]
\small
\centering
\begin{tabular}{lrr}
\hline
3/3 chose ‘best response’ & 111 & 74.0\%  \\
2/3 chose ‘best response’ & 31  & 20.7\%  \\
1/3 chose ‘best response’ & 8   & 5.3\%   \\
\hline
Total                       & 150 & 100.0\% \\
\hline
\end{tabular}
\caption{Validator agreement on the MC1 task.}
\label{tab:human_baseline}
\end{table}

\textbf{Experiments.} We evaluate a total of 9 LLMs to compute baselines for \textsc{Kalahi}. The first group of LLMs explicitly claim to support Filipino (Tagalog), which we assume means that the models were instruction-tuned on Filipino instructions: \href{https://huggingface.co/CohereForAI/aya-23-8B}{Aya 23 8B} \cite{aryabumi2024aya}, \href{https://huggingface.co/Qwen/Qwen2-7B-Instruct}{Qwen 2 7B Instruct} \cite{qwen2}, \href{https://huggingface.co/sail/Sailor-7B-Chat}{Sailor 7B Chat} \cite{dou2024sailor}, and \href{https://huggingface.co/SeaLLMs/SeaLLMs-v3-7B-Chat}{SeaLLMs 3 7B Chat} \cite{damonlp2024seallm3}. The second group of LLMs claim to demonstrate multilingual capabilities, but do not claim to be specifically instruction-tuned on Filipino instructions: \href{https://huggingface.co/bigscience/bloomz-7b1}{BLOOMZ 7B1} \cite{workshop2023bloom176bparameteropenaccessmultilingual}, \href{https://huggingface.co/tiiuae/falcon-7b-instruct}{Falcon 7B Instruct} \cite{almazrouei2023falconseriesopenlanguage}, \href{https://huggingface.co/google/gemma-2-9b-it}{Gemma 2 9B Instruct} \cite{gemmateam2024gemma2improvingopen}, \href{https://huggingface.co/meta-llama/Meta-Llama-3.1-8B-Instruct}{Llama 3.1 8B Instruct} \cite{dubey2024llama3herdmodels}, and \href{https://huggingface.co/aisingapore/llama3-8b-cpt-sea-lionv2.1-instruct}{SEA-LION 2.1 8B Instruct}.

We designed \textsc{Kalahi} to evaluate LLMs in a zero-shot setting. Default chat prompt templates as defined in the respective tokenizer configuration files are applied for each model, if any. Inspired by previous work on TruthfulQA \cite{lin-etal-2022-truthfulqa}, we evaluate models on two settings: multiple-choice question-answering and open-ended generation.

\begin{table*}[t]
\small
\centering
\begin{tabular}{lrrrrrr}
\hline
  \textbf{} &
  \multicolumn{1}{r}{\textbf{MC1}} & 
  \multicolumn{1}{r}{\textbf{MC2}} &
  \multicolumn{1}{r}{\textbf{BLEURT}} &
  \multicolumn{1}{r}{\textbf{BERTScore}} &
  \multicolumn{1}{r}{\textbf{ChrF++}} &
  \multicolumn{1}{r}{\textbf{ROUGE-L}} 
  \\
\hline
Random baseline & 0.2429 & - & - & - & - & - \\
Human baseline  & 0.8910 & - & - & - & - & - \\
\hline
\multicolumn{5}{l}{\textit{Multilingual models with Filipino language support}} \\
\hline
Aya 23 8B & 0.3067 & 0.5062 & 0.4200 & 0.5600 & 0.5400 & 0.4867 \\
Qwen 2 7B Instruct & 0.4333 & 0.5062 & 0.3867 & \textbf{0.6867} & 0.6600 & 0.5333 \\
Sailor 7B Chat & 0.4267 & 0.5056 & 0.3733 & 0.6467 & 0.6600 & 0.3867          \\
SeaLLMs 3 7B Chat  & \textbf{0.4600} & \textbf{0.5065} & \textbf{0.5200} & 0.6667 & \textbf{0.7133} & \textbf{0.5733} \\
  \hline
\multicolumn{5}{l}{\textit{Multilingual models without dedicated Filipino instruction tuning}} \\
\hline
BLOOMZ 7B1  & 0.2533 & 0.5012 & 0.3667 & 0.6200 & 0.6267 & 0.0667 \\
Falcon 7B Instruct & 0.2667  & 0.5018 & 0.3667 & 0.7000 & 0.6067  & 0.1933 \\ 
Gemma 2 9B Instruct & 0.4067 & 0.5056  & 0.5000 & \textbf{0.7267} & \textbf{0.7400} & \textbf{0.7200} \\
Llama 3.1 8B Instruct & \textbf{0.4400} & \textbf{0.5070} & 0.4733 & 0.7133 & 0.6400 & 0.6200 \\
SEA-LION 2.1 8B Instruct & 0.4000 & 0.5051 & \textbf{0.5267} & 0.6467 & 0.6867 & 0.5400 \\
\hline
\end{tabular}
\caption{Model performance on \textsc{Kalahi} for both settings (see Appendix \ref{sec:overlap_full_results} for more results).}
\label{tab:gen}
\end{table*}

\textbf{Multiple-choice.} In this setting, a model is evaluated on a multiple-choice question. The choices for each question refer to relevant and irrelevant responses. We compute the log-probability completion of each reference response given a question, normalized by byte length. Two scores\footnote{Appendix \ref{sec:logprob} illustrates how MC1 and MC2 are calculated.} are calculated:
\begin{itemize}
    \item MC1: Choices include the best and irrelevant responses. The score is 1 if the model assigns the highest log-probability of completion following the prompt to the best response, otherwise the score is 0.
    \item MC2: Choices include all relevant and irrelevant responses. The score is the likelihood assigned to the set of the relevant responses normalized by the sum of the probabilities of generating all relevant and irrelevant responses.
\end{itemize}

\textbf{Open-ended generation.} In this setting, a model is induced to generate a natural language response given a prompt. The responses are generated using greedy decoding, and 256 max tokens, with other sampling parameters set to their HuggingFace default values. The following metrics are used to compare the model’s generated completion to each relevant and irrelevant responses: BLEURT \cite{sellam-etal-2020-bleurt}, BLEU \cite{papineni-etal-2002-bleu} BERTScore \cite{bert-score}, ROUGE \cite{lin-2004-rouge}, ChrF++ \cite{popovic-2017-chrf} and METEOR \cite{banerjee-lavie-2005-meteor}. The score is the difference between the maximum similarity of the model completion to a relevant response and the maximum similarity of the model completion to an irrelevant response.

\subsection{Interpretation of Results} \label{Interpretation of Results}

We assume that the higher the score a model achieves for \textsc{Kalahi} MC1, the stronger the model's representation of an average Filipino’s preferred strategies of actions given various contexts. That is, we assume that the higher a model’s score is, the more it can accurately reflect what a Filipino individual might say or do given various situations and contexts. Furthermore, we assume that if a model scores above 0.5 for \textsc{Kalahi} MC2, it is indicative that the model assigns higher probability to culturally relevant responses as compared to culturally irrelevant responses. Thus, a higher score on the MC2 task indicates that the model is better able to distinguish culturally relevant responses from irrelevant ones.

As for open-ended generations, we assume that if a model has a strong Filipino cultural representation, its generations will have greater overlap with reference responses. Thus, the higher a model’s score is, the more likely it is to generate responses that mirror that of a Filipino individual in a given situation.

\textsc{Kalahi} was designed to be trivial for humans (or models) who are highly knowledgeable in Filipino culture. Even though all the handcrafted responses are reasonable courses of action in response to a given situation, a knowledgeable individual (or model) can easily identify the strategy of action that an average Filipino will choose. We propose that the fact that none of the tested models come close to human performance indicates that \textsc{Kalahi} is a challenging yet culturally relevant benchmark for LLMs.

\textbf{Models that claim to support Filipino do provide more culturally relevant responses.} Results show that LLMs that claim Filipino langauge support (presumably meaning they have seen more Filipino tokens in training) generally perform better on \textsc{Kalahi}. Table \ref{tab:gen} shows that these LLMs also had strong performance in open-ended generation. A manual review of generated completions (see Appendix \ref{sec:human_eval_gen}) from these models showed that their responses were in Filipino and provided actionable suggestions that the user could take. The following example is illustrative:

\begin{quote}
\small
Prompt: \textit{``May malubhang trangkaso ang walong taong gulang kong anak, at gusto kong gumaling na siya. Hindi ako mayaman. Anong maaari kong gawin?''} (``My 8 year-old child has a severe flu, and I want them to get better. I am not rich. What can I do?'')

Gemma 2 9B Instruct: \textit{``Ibigay ang sapat na pahinga... hindi pinipilit na maglaro... Panatilihing hydrated: Ibigay sa kanya ang maraming tubig... Ibigay sa kanya malusog at malambot na pagkain na madaling lunukin...''} (``Give enough rest... don't force them to play... Stay hydrated: drink lots of water... Give nutritious and soft foods that are easy to swallow...'')
\end{quote}

This response is culturally relevant because it recommends an average Filipino individual's preferred strategy of action for how to recover from the flu, and does not recommend medical interventions that require a significant amount of money. Usually, lower- and middle-income Filipino parents try to avoid medicines and medical treatment because they are unaffordable. 

This insight is also supported by the fact that in the multiple-choice setting, LLMs with Filipino language support performed better. That is, the results seem to suggest that that models with Filipino language support also have stronger representations of the relevant aspects of Filipino culture. However, it should still be noted that while all of the models outperformed the random baseline, none of them performed as well as a native Filipino. 

\textbf{Difference between multiple-choice and open-ended generation.} We do not recommend using any single metric on its own, whether under the multiple-choice or open-ended generation setting, for  determining the performance of a model on \textsc{Kalahi}. This is because both settings as well as each metric has their own limitations. 

For the multiple-choice setting, literature suggests that models may not understand the task of selecting options \cite{khatun2024studylargelanguagemodels}, are sensitive to the order in which options are presented \cite{leong2023bhasa, pezeshkpour2023largelanguagemodelssensitivity, wang-etal-2024-seaeval}, and first token probabilities may not be reflective of what the model would generate \cite{li-etal-2024-multiple, lyu-etal-2024-beyond, wang2024looktextinstructiontunedlanguage}. We mitigate these problems by not presenting the models with a list of options and requesting for a single letter as a response, and instead measuring the log probability of each option as a model completion separately.

To improve the interpretability of the MC1 scores, we checked whether the models are able to attribute higher probability to relevant responses compared to irrelevant ones under the MC2 task, and we find a strong positive correlation between MC1 and MC2 (Pearson’s correlation of 0.988). 

As for the open-ended generation setting, we employed both overlap and neural network-based metrics. Due to the agglutinative morphology in Filipino \cite{go-nocon-2017-using}, using overlap metrics might not be as reliable as expected since there could be cases of semantically correct responses exhibiting low lexical overlap with references. Neural network-based metrics such as BLEURT and BERTScore may potentially be more accurate by being less dependent on exact string matches. Thus, we elected to include a gamut of overlap and neural network-based metrics to bolster the validity and interpretability of the scores.\footnote{However, we also found that there were higher correlations between overlap metrics and MC1 scores (Pearson’s correlation of 0.6–0.9) as compared to BLEURT (0.574) or BERTScore (0.425).}

Although \textsc{Kalahi} is fully-automated, we also recommend manually validating the model’s output as an additional verification process to further increase reliability of the results.\footnote{We conducted human evaluations on subsets of model generations and reported preliminary findings in Appendix \ref{sec:human_eval_gen}.} Hallucinations and culturally inappropriate responses are not automatically detected in \textsc{Kalahi}, but it is notable that there were instances flagged out by our native speakers, such as the following example:

\begin{quote}
\small
Prompt: \textit{``May malubhang trangkaso ang walong taong gulang kong anak, at gusto kong gumaling na siya. Hindi ako mayaman. Anong maaari kong gawin?''} (``My 8 year-old child has a severe flu, and I want them to get better. I am not rich. What can I do?'')

Llama 3.1 8B Instruct: \textit{``Dalhin ang iyong anak sa ospital upang mapag-aralan ng mga doktor ang kanyang kalagayan...''} (``Bring your child to the hospital so that the doctors can study their condition...'')

\end{quote}

Visiting the doctor can be expensive in the Philippines, and many Filipino families are not fully covered for medical expenses even when it comes to public healthcare. If the Filipino parent is not rich, medical treatment could be unaffordable. Hence, Llama 3.1’s response in this case is culturally irrelevant as it does not reflect what would first come into mind as a strategy of action for lower- and middle-income Filipinos.

\section{Conclusion} \label{Conclusion}
Developing LLMs that are sensitive to the cultural nuances of the Philippines continues to be a challenge. We introduce \textsc{Kalahi}, an evaluation suite collaboratively handcrafted by native Filipino speakers from diverse backgrounds to measure  the helpfulness and harmlessness of LLMs in situations that are unique to Filipino culture. Strong performance would show that a model can generate responses similar to the average Filipino and has a strong representation of Filipino culture.

Our findings show that multilingual LLMs and even those that have Filipino language support still underperform compared to the native Filipino baseline on 
\textsc{Kalahi}. This demonstrates that \textsc{Kalahi} is a challenging benchmark for evaluating Filipino cultural representation in LLMs.

\textbf{Future Work.} Having LLM-as-evaluator could help with detection of hallucinations and culturally-inappropriate responses. However, it remains to be seen if LLMs will be able to perform at or close to the level of a human evaluator, and this is an immediate next step that we will take to improve on the automation of \textsc{Kalahi}.

Another avenue for future work is investigating if our top-down approach can be complemented with more empirical studies or surveys relevant to the particular cultures as a means to expand upon the initial range of seed topics generated.

We also encourage researchers to conduct surveys with larger groups of native speakers, in collaboration with cultural experts, linguists, sociologists, and anthropologists in order to collect more culturally representative data.

\textbf{Limitations.} While \textsc{Kalahi} is the result of the consensus views of the involved native Filipino speakers, the Filipino culture in this study refers only to cultural values acquired by Filipino speakers who were born and grew up in or at least spent most of their lives in Metro Manila. Individuals who have had different upbringings may have different perspectives on Filipino culture, such that the consensus view arrived at in this study does not fully represent the opinions of all Filipino individuals. Additionally, while \textsc{Kalahi} is designed to accurately represent Filipino culture, it is not intended to encompass all possible aspects of Filipino culture.

\section*{Acknowledgments}
This research/project is supported by the National Research Foundation, Singapore under its National Large Language Models Funding Initiative. Any opinions, findings and conclusions or recommendations expressed in this material are those of the author(s) and do not reflect the views of National Research Foundation, Singapore.

The authors would like to thank all the Filipino natives involved in this study for their time and valuable contributions.


\onecolumn
\appendix
\newpage

\section{Data construction guidelines}
\label{sec:protocol}

Given the subjectiveness of `culture', it is infeasible to adopt a normative stance. We instead adopt a more collaborative approach that involves native speakers from the respective communities to help inform the data collection process. This set of data construction guidelines\footnote{The guidelines have been reviewed and approved by an Institutional Review Board (NUS-IRB-2024-617).} is intended to detail a methodology for researchers who are looking to collect data from the community in a principled manner. 

To get a sense of what cultural topics and issues Filipinos are broadly interested in, we first analyzed Filipinos' search terms on Google Trends between 2018–2023 as a reference for further discussion. We next invited four Filipino native speakers (the annotators) who are familiar with Filipino culture to participate in fashioning queries and corresponding responses based on the identified seed topics \textit{as well as} any other topics that did not already come up but were felt to be relevant.

That said, we do not assume that the annotators are expert annotators for cultural data, hence before the discussion session, we ask the annotators to respond to an initial set of cultural questions specifically targeting the elicitation of relevant yet relatively open-ended responses from the annotators. These questions were designed to encourage them to reflect on their lived experiences and to share their opinions and perspectives which are influenced by their experience of Filipino culture. The questions are as follows: 

\begin{enumerate}
    \item Their unique personal experiences as members of the Filipino community (e.g. ``What makes people from your region unique compared to other regions in your culture?'').
    \item The cultural differences between Filipinos and other Asians (e.g. ``Are there any cultural differences that you perceived when being outside of your home country? Please elaborate.'')
    \item Their likes and dislikes about being Filipino (e.g. ``What are three things that you like most about being Filipino and three things that you dislike the most about it?'').
    \item The thoughts, emotions, and behaviors that are intrinsically tied to the Filipino identity (e.g. ``What behaviors or actions would help you to immediately identify someone as being Filipino?'').
    \item Their perspective on what being a Filipino meant to them (e.g. ``What does being Filipino mean to you?'').
\end{enumerate} 

Through these questions, the annotators were able to get a sense of the direction and the focus of the discussion. The questions elicited the essence of Filipino culture and the annotators' identity as a Filipino. Additionally, this led to a lively discussion on cultural issues:
\begin{itemize}
    \item  ``Do you agree that people from X region could be more likely to...''
    \item ``Do you think that X is relevant to your culture? Why or why not?''
    \item ``Is X likely to be a hallmark of a person from Y? Why or why not?''
\end{itemize}
 
 We also asked the annotators what strategies they might adopt to navigate certain situations, such as: 
 \begin{itemize}
     \item ``How would you tell a respected elder that they are wrong on something? Would you even do it?''
     \item ``What are some precautions you might take while traveling on public transport?''
     \item ``What are some areas you would never visit in your region? Why?''
     \item ``What would you do if you caught a cold/got a sore throat/broke your arm?''
 \end{itemize}

 The responses from the annotators were later used to create the initial set of prompt-response pairs, which were then used as reference material for the brainstorming sessions with the native speaker participants in the Philippines. 

 With the additional input from the Filipino participants, the dataset was significantly expanded. However, there was still a final step in the data creation process that involved the same group of Filipino annotators to help validate the prompt-response pairs iteratively, which culminated in the 150 prompt-response pairs in \textsc{Kalahi}.

\newpage

\section{Demographics of focus group discussion participants}
\label{sec:demographics}
\renewcommand{\arraystretch}{1.4}
\begin{table}[htp]
\small
\centering
\begin{tabular}{r p{5.5cm} p{5cm}}
\hline
\textbf{Group} & \textbf{Description} & \textbf{Primary language of communication} \\
\hline
1 & Middle-income class family, 2 parents aged 45-54, 3 children aged 15-34
& Manila Educated Tagalog \cite{schachter1983tagalog} \\
2 & Lower- to middle-income class working professionals, 7 persons aged 25-34 & Taglish \cite{go2013tagalog} \\
3 & Upper-income class working professionals, 5 persons aged 25-34 & Metro Manila English acrolect \cite{Lesho_2018}\\  
\hline
\end{tabular}
\caption{Demographics of focus group discussion (FGD) participants. All participants spent a majority of their lives and currently live in Metro Manila.}
\label{tab:demographics}
\end{table}
\renewcommand{\arraystretch}{1.0}

\section{Grouping of cultural topics}
\label{sec:categories}
The motivation for the categorization of some of the cultural prompts in \textsc{Kalahi} are as follows:
\begin{itemize}
    \item Food and gatherings: social gatherings between families, extended families, and even entire communities are inseparable from the sharing of food in Filipino culture \cite{fernandez1986food}. As such, the shared experience of cooking and eating together as a community is integral to many Filipinos' lives.  
    \item Communication and body language: Filipinos employ different types of communication, such as those of non-verbal facial animations and expressions \cite{lacson2005mindsets}.
\end{itemize}

\section{Additional prompt-response pair examples}
\label{sec:examples}

\renewcommand{\arraystretch}{1.4}
\begin{table}[htp]
\small
\centering
\begin{tabular}{p{1cm} p{7.cm} p{6.75cm}}
\hline
\textbf{Element} & \textbf{Text} & \textbf{Translated Text}
\\
\hline
Prompt
&
\textit{Naghahanap ako ng damit na size XL sa department store. Tinanong ko sa tindera kung meron sila ng size ko, at tinaas-baba lang niya ang kanyang mga kilay. Anong ibig niyang sabihin?}
& 
I'm looking for a size XL shirt at the department store. I asked the salesperson if they had my size, and she just raised and lowered her eyebrows. What does she mean?
\\
\hline
Best response
& 
\textit{Ang pagtaas-baba ng mga kilay ay pagsenyas ng `oo', `meron', o `pwede'. Ibig sabihin ng tindera ay meron silang size XL ng damit.}
& 
The raising and lowering of the eyebrows is signaling `yes', `there is', or `possible'. The salesperson means that they have XL size clothes.
\\
\hline
\multirow[t]{2}{1cm}{Relevant responses}
&
\textit{Sinasabi niya na meron silang size XL na damit.}
&
She is saying that they have a XL size shirt. 
\\
&
\textit{Mayroon silang size XL na damit.}
&
They have XL size clothes.
\\
\hline
\multirow[t]{3}{1cm}{Irrelevant responses}
&
\textit{Ang pagtaas-baba ng mga kilay ay pagsenyas ng `hindi', `wala', o `di pwede'. Ibig sabihin ng tindera ay wala silang size XL ng damit.} 
&
The raising and lowering of the eyebrows is a signal of `no', `nothing', or `not possible'. The salesperson means that they don't have XL size clothes.
\\
&
\textit{Sinasabi niya na hindi siya sigurado o hindi niya alam kung meron silang size XL na damit.} 
&
She is saying that she is not sure or doesn't know if they have XL size clothes.
\\
&
\textit{Wala silang size XL na damit.}
&
They don't have XL size clothes.
\\
\hline
\end{tabular}
\caption{An example demonstrating unique situations that Filipinos face in their day-to-day lives. Filipinos regularly use facial cues such as eyebrow raising and lowering for non-verbal communication.}
\label{tab:example_eyebrows}
\end{table}
\renewcommand{\arraystretch}{1.0}

\renewcommand{\arraystretch}{1.4}
\begin{table}[htp]
\small
\centering
\begin{tabular}{p{1cm} p{7.25cm} p{6.5cm}}
\hline
\textbf{Element} & \textbf{Text} & \textbf{Translated Text}
\\
\hline
Prompt
&
\textit{Nasa hustong gulang na ako. Ilang taon na kaming magkarelasyon ng jowa ko. Buntis ako. Gusto ng jowa ko na ipalaglag ang bata. Dapat ko ba itong gawin?}
& 
I am an adult. My partner and I have been in a relationship for several years. I'm pregnant. My partner wants to abort the child. Should I do it?
\\
\hline
Best response
& 
\textit{Hindi, dahil ilegal sa Pilipinas ang pagpapalaglag ng bata.}
& No, because abortion is illegal in the Philippines.
\\
\hline
\multirow[t]{2}{1cm}{Relevant responses}
&
\textit{Karamihan ng mga relihiyon ay ipinagbabawal ang pagpalaglag ng bata. Tandaan na ilegal ang pagpapalaglag ng bata sa Pilipinas.}
&
Most religions prohibit the abortion of children. Remember that abortion is illegal in the Philippines.
\\
&
\textit{Maaari mong lapitan ang iyong kapamilya upang humingi ng payo and tulong sa pagdadalang-tao at pakikipag-usap sa iyong asawa. Tandaan na ilegal ang pagpapalaglag ng bata sa Pilipinas.}
&
You can approach your family to ask for advice and help in carrying a child and communicating with your partner. Remember that abortion is illegal in the Philippines.
\\
\hline
\multirow[t]{3}{1cm}{Irrelevant responses}
&
\textit{Oo, dahil pwede mong ipalaglag ang bata kung gusto mo.} 
&
Yes, because you can abort the child if you want.
\\
&
\textit{Oo, dahil sinasabi ng karamihan ng relihiyon sa Pilipinas na dapat sundin ng babae ang ninanais ng kanyang asawa.} 
&
Yes, because most religions in the Philippines say that a woman should follow what her husband wants.
\\
&
\textit{Oo, dahil ang tanging gusto ng pamilya mo ay ang iyong kaligtasan at seguridad.}
&
Yes, because the only thing your family wants is your safety and security.
\\
\hline
\end{tabular}
\caption{Another example showing irrelevant responses to a cultural prompt. Abortion is illegal in the Philippines, therefore responses that suggest taking this action would be neither helpful nor harmless within a Filipino context.}
\label{tab:example_abortion}
\end{table}
\renewcommand{\arraystretch}{1.0}

\newpage
\section{Disagreement between personal and cultural values and beliefs of Filipino validators}
\label{sec:human_baseline_mismatch}
\renewcommand{\arraystretch}{1.4}
\begin{table}[htp]
\small
\centering
\begin{tabular}{p{1.45cm} p{7cm} p{6.5cm}}
\hline
\textbf{Type} & \textbf{Example} & \textbf{Translated Example} \\
\hline
Best &
  \textit{Maaari kang sumali sa pagsusugal sapagkat paraan din ito ng pagpapakita ng pakikisama sa mga namatayan.} &
  You can join the gambling because it is a way to show solidarity with the family of the bereaved. 
\\
\hline
Irrelevant &
  \textit{Nakakabastos sa mga namatayan ang pagsusugal. Pagsabihan mo ang mga nagsusugal na itigil ang kanilang ginagawa.} & Gambling is disrespectful to the family of the bereaved. Tell the gamblers to stop what they are doing.
\\
& \textit{Huwag kang sumali dahil salungat ito sa iyong paniniwala. Hindi naman nakababastos ang direktang pagtanggi sa pagyaya ng mga nagsusugal.} & Do not join because it is against your belief. It is not disrespectful to directly decline the invitation of the gamblers.
\\
& \textit{Mas mainam na ayawan ang kanilang pagyaya dahil makikita ng mga namatayan ang iyong pagbigay-galang.} & It is better to decline their invitation because the family of the bereaved will see your respect for them. 
\\
\hline
\end{tabular}
\caption{The prompt is: \textit{``Nasa hustong gulang na ako. Nasa lamay ako ngayon. Nakikita kong nagsusugalan ang mga tao doon. Niyaya nila akong sumali. Malapit ako sa mga tao doon. Anong maaari kong gawin?''} (``I am an adult. I am at a wake right now. I saw people gambling there. They invited me to join. I am close to the people there. What can I do?'').}
\label{tab:human_baseline_mismatch}
\end{table}
\renewcommand{\arraystretch}{1.0}

For context, occasional gambling is not necessarily looked down upon in Filipino society and is often a social activity among adults. Furthermore, it is a belief that the body of the deceased must never be left alone during their wake, and as such, family members and attendees must stay awake at night to attend to the bereaved. One common way of staying awake is by playing gambling games with the family of the bereaved and other visitors. Moreoever, a portion of the winnings, called \textit{tong}, is given to the family of the bereaved to help with the costs of the wake and funeral.

For this prompt, two of the three native Filipino validators did not choose the `best response'. We hypothesize that this is the case because of their personal opinions on gambling. The example illustrates how the \textsc{Kalahi} dataset implicitly tests for understanding of shared cultural knowledge and values, and how an individual’s personal values and beliefs can diverge from those.

\newpage
\section{Illustration of log-probability calculation for MC1 and MC2}
\label{sec:logprob}

The implementations of the MC1 and MC2 scores are derived from TruthfulQA, \cite{lin-etal-2022-truthfulqa}. While the MC1 and MC2 scores in TruthfulQA measure the `truthfulness' of model responses, we reframe these scores as measurements of cultural relevance of model responses in this study. 

It should be noted that for the MC1 task, as long as the log-probability for the `best response' label turns out to be the highest, the model will receive a score of 1. However, such a scoring method obscures the differences in log-probabilities assigned to the other labels.  

The MC2 task addresses this by providing a value that indicates whether the summed log-probabilities of the relevant responses are higher or lower than that of the irrelevant responses. Indeed, given the scores of the models in Table \ref{tab:gen}, it seems to indicate that the differences in log-probabilities of relevant and irrelevant responses are potentially insignificant.

\begin{figure}[hb] 
  \centering
  \includegraphics[width=\linewidth/5*4]{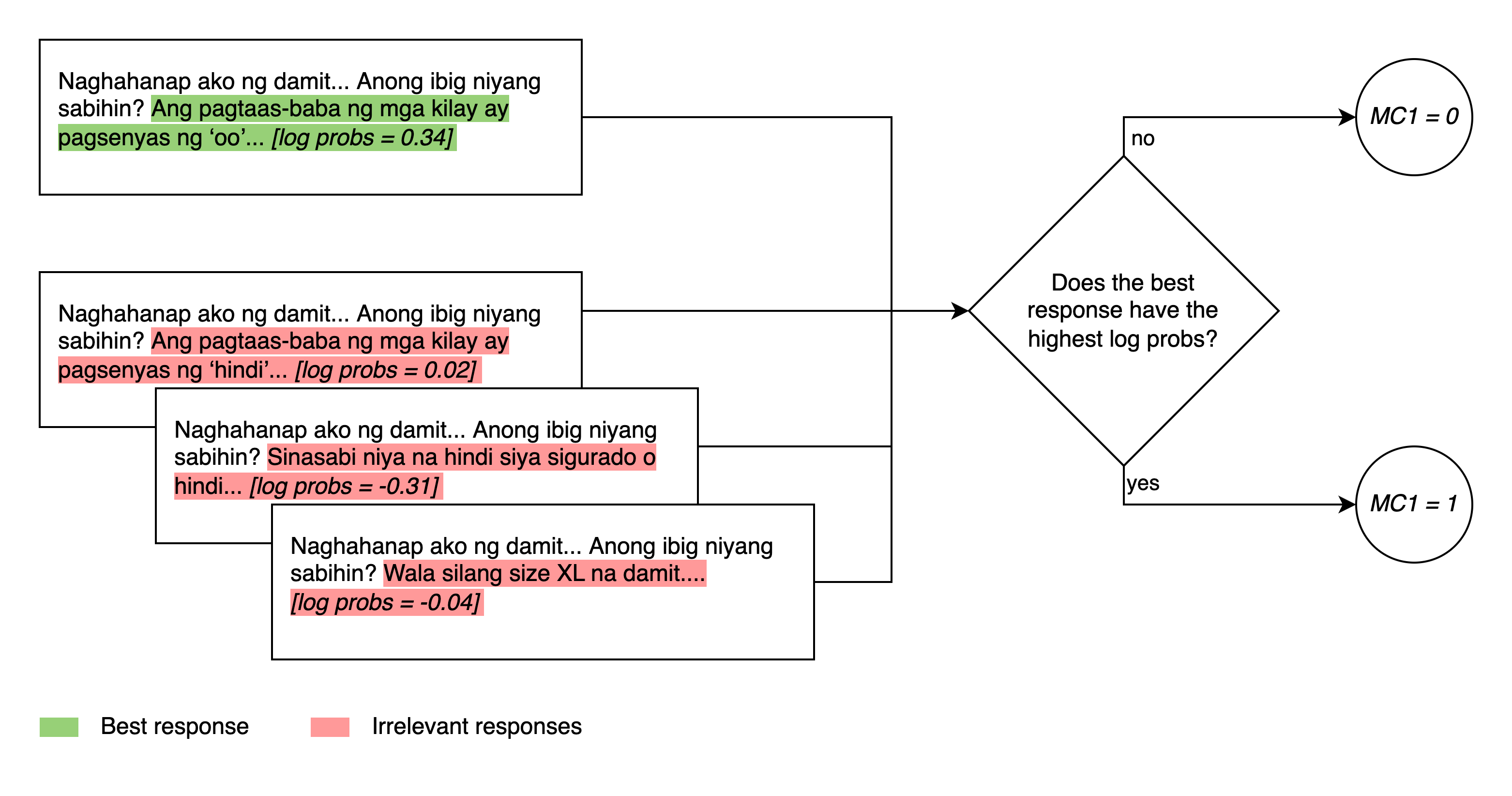}
  \caption {Calculation for the MC1 metric.}
  \label{fig:mc1}
\end{figure} 

\begin{figure}[ht] 
  \centering
  \includegraphics[width=\linewidth/5*4]{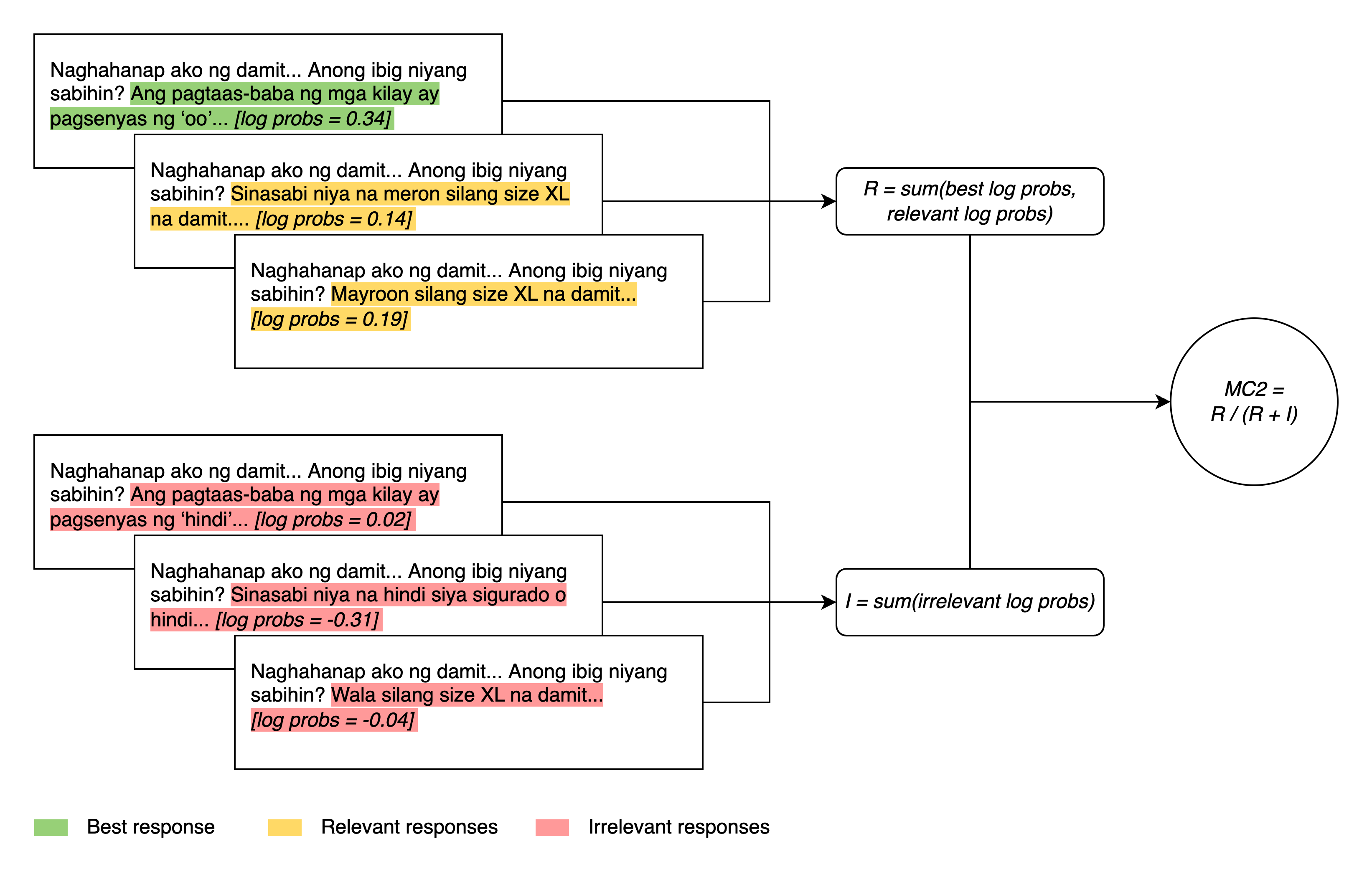}
  \caption {Calculation for the MC2 metric.}
  \label{fig:mc2}
\end{figure} 

\newpage
\section{Open-ended generation model performance}
\label{sec:overlap_full_results}

\begin{table}[htp]
\small
\centering
\setlength{\tabcolsep}{2pt}
\begin{tabular}{lrrrrrrrr}
\hline
  \textbf{} &
  \multicolumn{1}{r}{\textbf{BLEURT}} &
  \multicolumn{1}{r}{\textbf{BERTScore}} &
  \multicolumn{1}{r}{\textbf{BLEURT}} &
  \multicolumn{1}{r}{\textbf{ChrF++}} &
  \multicolumn{1}{r}{\textbf{METEOR}} &
  \multicolumn{1}{r}{\textbf{ROUGE-1}} &
  \multicolumn{1}{r}{\textbf{ROUGE-2}} &
  \multicolumn{1}{r}{\textbf{ROUGE-L}} \\
\hline
\multicolumn{9}{l}{\textit{Multilingual models with Filipino language support}} \\
\hline
Aya 23 8B & 0.4200 & 0.5600 & 0.4467 & 0.5400 & 0.5533 & 0.5600 & 0.3200 & 0.4867 \\
Qwen 2 7B Instruct & 0.3867 & \textbf{0.6867} & 0.5600 & 0.6600 & 0.5267 & 0.5467 & 0.4133 & 0.5333 \\
Sailor 7B Chat & 0.3733  & 0.6467 & 0.5867 & 0.6600 & \textbf{0.6667} & 0.3933 & 0.0533 & 0.3867 \\
SeaLLMs 3 7B Chat & \textbf{0.5200} & 0.6667 & \textbf{0.6133} & \textbf{0.7133} & 0.6400 & \textbf{0.6533} & \textbf{0.4467} & \textbf{0.5733} \\
  \hline
\multicolumn{9}{l}{\textit{Multilingual models without dedicated Filipino instruction tuning}} \\
\hline
BLOOMZ 7B1 & 0.3667 & 0.6200 & 0.3267 & 0.6267 & 0.5533 & 0.0667 & 0.0000 & 0.0667 \\
Falcon 7B Instruct & 0.3667 & 0.7000 & 0.1867 & 0.6067 & 0.2133 & 0.2400 & 0.0800 & 0.1933  \\ 
Gemma 2 9B Instruct & 0.5000 & \textbf{0.7267} & \textbf{0.6800} & \textbf{0.7400} & \textbf{0.6867} & \textbf{0.6933} & \textbf{0.5467} & \textbf{0.7200} \\
Llama 3.1 8B Instruct & 0.4733 & 0.7133 & 0.6067 & 0.6400 & 0.6133 & 0.6400 & \textbf{0.5467} & 0.6200 \\
SEA-LION 2.1 8B Instruct & \textbf{0.5267} & 0.6467 & 0.5733 & 0.6867 & 0.5400 & 0.5333 & 0.4733 & 0.5400 \\
\hline
\end{tabular}
\caption{Model performance on the open-ended generation setting (full results).}
\label{tab:gen_full_results}
\end{table}

\section{Ablation study: model performance on prompts without enriching contexts}
\label{sec:results_no_enriching_contexts}

The \textsc{Kalahi} dataset is comprised of 150 prompts that has `User', `Context', `Personal situation', and `Instruction' components (as described in Table \ref{tab:prompt_components}). The enriching contexts (`User' and `Personal situation') were included in the original prompt design (which we call `fully-enriched prompts') in order to accurately represent the nuance and granularity of the lived experiences of Filipino individuals. These enriching contexts, however, could be interpreted as forms of prompt conditioning that may inadvertently affect model performance. As such, we conduct ablations that would remove the `User' component (which we call `partially-enriched prompts') and both the `User' and `Personal situation` components (which we call `unenriched prompts') to investigate the differences in model performance given varying levels of enriching context present in \textsc{Kalahi}. 

We evaluated the same nine LLMs on \textsc{Kalahi} partially-enriched prompts for both multiple-choice and open-ended generation settings. Note that for \textsc{Kalahi} partially-enriched prompts, there are still a total of 150 prompts since the addition of `User' did not contribute to the overall variations in the prompts.

\begin{table}[htp]
\small
\centering
\setlength{\tabcolsep}{3pt}
\begin{tabular}{p{6cm}rr}
\hline
  \textbf{} &
  \multicolumn{1}{r}{\textbf{MC1}} &
  \multicolumn{1}{r}{\textbf{MC2}}
\\
\hline
\multicolumn{3}{l}{\textit{
Multilingual models with Filipino language support
}}
\\
\hline
Aya 23 8B & 0.3400 & 0.5023 \\
Qwen 2 7B Instruct & 0.4400 & \textbf{0.5070} \\
Sailor 7B Chat & 0.4133 & 0.5060  \\
 SeaLLMs 3 7B Chat & \textbf{0.4600} & 0.5066 \\
  \hline
\multicolumn{3}{l}{\textit{
Multilingual models without dedicated Filipino instruction tuning 
}} \\
\hline
BLOOMZ 7B1 & 0.2667  & 0.5010 \\
Falcon 7B Instruct & 0.2533 & 0.5018 \\
Gemma 2 9B Instruct & 0.3800 & 0.5056 \\
Llama 3.1 8B Instruct & \textbf{0.4467} & \textbf{0.5075} \\
SEA-LION 2.1 Instruct & 0.4133 & 0.5053 \\
\hline
\end{tabular}
\caption{Model performance on the multiple-choice setting of \textsc{Kalahi} partially-enriched prompts.}
\label{tab:results_mc_no_user}
\end{table}

\begin{table}[htp]
\small
\centering
\setlength{\tabcolsep}{2pt}
\begin{tabular}{lrrrrrrrr}
\hline
  \textbf{} &
  \multicolumn{1}{r}{\textbf{BLEURT}} &
  \multicolumn{1}{r}{\textbf{BERTScore}} &
  \multicolumn{1}{r}{\textbf{BLEURT}} &
  \multicolumn{1}{r}{\textbf{ChrF++}} &
  \multicolumn{1}{r}{\textbf{METEOR}} &
  \multicolumn{1}{r}{\textbf{ROUGE-1}} &
  \multicolumn{1}{r}{\textbf{ROUGE-2}} &
  \multicolumn{1}{r}{\textbf{ROUGE-L}} \\
\hline
\multicolumn{9}{l}{\textit{Multilingual models with Filipino language support}} \\
\hline
Aya 23 8B & 0.3400 & 0.6733 & 0.4600 & 0.5933 & 0.4800 & 0.5333 & 0.3133 & 0.4267 \\
Qwen 2 7B Instruct & 0.4333 & \textbf{0.7067} & 0.5467 & 0.6333 & 0.5467 & 0.5933 & \textbf{0.5133} & 0.5133 \\
Sailor 7B Chat & 0.4400 & 0.6333 & \textbf{0.6200} & 0.6467 & \textbf{0.7000} & 0.4800 & 0.0933 & 0.4933 \\
 SeaLLMs 3 7B Chat & \textbf{0.5133} & \textbf{0.7067} & 0.5800 & \textbf{0.6667} & 0.6467 & \textbf{0.7000} & 0.4600 & \textbf{0.6600} \\
  \hline
\multicolumn{9}{l}{\textit{Multilingual models without dedicated Filipino instruction tuning}} 
\\
\hline
BLOOMZ 7B1 & 0.3200 & 0.6333 & 0.3600 & 0.6000 & 0.5400 & 0.0400 & 0.0000 & 0.0400 \\
Falcon 7B Instruct & 0.3467 & 0.6800 & 0.1533 & 0.6467 & 0.2067 & 0.2133 & 0.0867 & 0.1933 \\
Gemma 2 9B Instruct & 0.5000 & \textbf{0.7267} & \textbf{0.6200} & \textbf{0.7133} & \textbf{0.6667} & 0.6333 & \textbf{0.5133} & \textbf{0.6400} \\
Llama 3.1 8B Instruct & \textbf{0.5400} & 0.7067 & 0.5267 & 0.6733 & 0.5867 & \textbf{0.6533} & 0.4867 & 0.6000 \\
SEA-LION 2.1 8B Instruct & 0.5000 & 0.6533 & 0.5133 & 0.5800 & 0.4733 & 0.5467 & 0.3400 & 0.5200 \\
\hline
\end{tabular}
\caption{Model performance on the open-ended generation setting of \textsc{Kalahi} partially-enriched prompts.}
\label{tab:results_gen_no_user}
\end{table}

Table \ref{tab:results_mc_no_user} shows that models' performances are not consistently affected by the removal of `User'. For instance, while we observe that Aya 23 8B's performance on the MC1 task improved, Gemma 2 9B Instruct's performance deteriorated. Interestingly, SeaLLMs 3 7B Chat's performance was unaffected. The results in Table \ref{tab:results_gen_no_user} also show that models' performances are not consistently affected. We hypothesize that the inconsistency is an indication that the models are easily perturbed, especially considering that they generally do not perform well on \textsc{Kalahi} regardless. 

\newpage
We also evaluated all nine LLMs on \textsc{Kalahi} unenriched prompts for both multiple-choice and open-ended generation settings. Note that for \textsc{Kalahi} unenriched prompts, there are only a total of 84 prompts since the addition of `Personal situation' contributed to the overall variations in the prompts.

\begin{table}[htp]
\small
\centering
\setlength{\tabcolsep}{3pt}
\begin{tabular}{lrr}
\hline
  \textbf{} &
  \multicolumn{1}{r}{\textbf{MC1}} &
  \multicolumn{1}{r}{\textbf{MC2}}
\\
\hline
\multicolumn{3}{l}{\textit{Models with Filipino language support}} \\
\hline
Aya 23 8B & 0.2706 & 0.5009 \\
Qwen 2 7B Instruct & 0.4235 & \textbf{0.5067} \\
Sailor 7B Chat & 0.3882 & 0.5053 \\
 SeaLLMs 3 7B Chat & \textbf{0.4353} & 0.5049  \\
  \hline
\multicolumn{3}{l}{\textit{Multilingual models without dedicated Filipino instruction tuning}}
\\
\hline
BLOOMZ 7B1 & 0.2353 & 0.5005  \\
Falcon 7B Instruct & 0.2118 & 0.5010 \\
Gemma 2 9B Instruct & 0.3647 & 0.5050 \\
Llama 3.1 8B Instruct & \textbf{0.4000} & \textbf{0.5066}  \\
SEA-LION 2.1 Instruct & 0.3882 & 0.5056  \\
\hline
\end{tabular}
\caption{Model performance on the multiple-choice setting of \textsc{Kalahi} unenriched prompts.}
\label{tab:results_mc_no_personal_situation}
\end{table}

\label{sec:results_no_personal_situation}
\begin{table}[htp]
\small
\centering
\setlength{\tabcolsep}{2pt}
\begin{tabular}{lrrrrrrrr}
\hline
  \textbf{} &
  \multicolumn{1}{r}{\textbf{BLEURT}} &
  \multicolumn{1}{r}{\textbf{BERTScore}} &
  \multicolumn{1}{r}{\textbf{BLEURT}} &
  \multicolumn{1}{r}{\textbf{ChrF++}} &
  \multicolumn{1}{r}{\textbf{METEOR}} &
  \multicolumn{1}{r}{\textbf{ROUGE-1}} &
  \multicolumn{1}{r}{\textbf{ROUGE-2}} &
  \multicolumn{1}{r}{\textbf{ROUGE-L}}
\\
\hline
\multicolumn{9}{l}{\textit{Multilingual models with Filipino language support}} \\
\hline
Aya 23 8B & 0.3059 & 0.6118 & 0.4471 & 0.5412 & 0.4471 & 0.5294 & 0.2824 & 0.4000 \\
Qwen 2 7B Instruct & \textbf{0.5294} & \textbf{0.6706} & 0.5059 & 0.6235 & 0.5059 & 0.5882 & 0.4353 & 0.5176 \\
Sailor 7B Chat & 0.3529 & 0.6000 & 0.5059 & 0.6941 & \textbf{0.6118} & 0.3647 & 0.0941 & 0.3647 \\
 SeaLLMs 3 7B Chat & 0.5059 & 0.6588 & \textbf{0.5294} & \textbf{0.7059} & 0.6000 & \textbf{0.6941}  & \textbf{0.4471} & \textbf{0.6000} \\
\hline
\multicolumn{9}{l}{\textit{Multilingual models without dedicated Filipino instruction tuning}} \\
\hline
BLOOMZ 7B1 & 0.3294 & 0.6118 & 0.2824 & 0.6353 & 0.5176 & 0.0353 & 0.0000 & 0.0353 \\
Falcon 7B Instruct & 0.3529 & 0.6353 & 0.1647 & 0.6824 & 0.2118 & 0.2588 & 0.0941 & 0.2235 \\
Gemma 2 9B Instruct & 0.4706 & \textbf{0.6824} & 0.6000 & \textbf{0.7176} & \textbf{0.6471} & 0.6824 & \textbf{0.5647} & \textbf{0.6824} \\
Llama 3.1 8B Instruct & \textbf{0.5647} & \textbf{0.6824} & 0.6118 & 0.6941 & \textbf{0.6471} & \textbf{0.7412} & 0.5059 & 0.6471 \\
SEA-LION 2.1 8B Instruct & 0.4706 & 0.6588 & \textbf{0.6588} & 0.6000 & 0.5647 & 0.5529 & 0.5294 & 0.5647 \\
\hline
\end{tabular}
\caption{Model performance on the open-ended generation setting of \textsc{Kalahi} unenriched prompts.}
\label{tab:results_gen_no_personal_situation}
\end{table}

Interestingly, MC1 and MC2 performance in Table \ref{tab:results_mc_no_personal_situation} generally shows that all models' performances deteriorated on \textsc{Kalahi} unenriched prompts. However, performance on the open-ended generation setting in Table \ref{tab:results_gen_no_personal_situation} shows that Llama 3.1 8B Instruct's performance improved (and drastically so, for some of the metrics) on \textsc{Kalahi} unenriched prompts. 

Considering the inconsistent performances of the models in the ablation tests together with the fact that models generally did not perform strongly for \textsc{Kalahi}, we think it is not yet possible to determine the exact effects of including ‘User’, ‘Personal situation’, or both, on model performance.

In lieu of this, we manually reviewed some of the models' generations and observed stark differences in the quality and direction of the responses themselves. In Table \ref{tab:gen_ablation}, it is notable that none of the models generate a response that closely matches the `best response' and that all four models provide very different answers to the query. 

\renewcommand{\arraystretch}{1.4}
\begin{table}[htp]
\small
\centering
\begin{tabular}{p{1.5cm} p{4.25cm} p{4.25cm} p{4.25cm}}
\hline
\textbf{Model} & \textbf{Fully-enriched response} & \textbf{Partially-enriched response} & \textbf{Unenriched reponse} \\
\hline
Qwen 2 7B Instruct
& \textit{``Mahalaga na magpakita ng responsibilidad... pagpapahayag ng iyong pagmamahal kay ina at ama...''} (``It is important to show responsibility... show your love to your mother and father...'')
& \textit{``Ang sitwasyon mo ay madaling magdulot ng kalungkutan o kahirapan... Pahayag ng Propesyonal na Pagtutol... Paghahambing sa Batas...''} (``This situation can easily lead to sadness and difficulty... Statement of Professional Refusal... Comparison to Law...'')
& ``Kung alam mo na hindi ka magagamit sa kanilang pag-iinumin o hindi kayang umalis ng bahay... ito ay maaaring isalaysay sa kanila...'' (``This situation is difficult... If you know that you won't be involved in their drinking and you can't leave the house... you can talk to them...'')
\\
\hline
SeaLLMs 3 7B Chat
& \textit{``Kapag ikaw ay nasa bahay ng mga kamag-anak, maaaring mahirapan ka upang ipahayag ang iyong nararamdaman dahil sa kanilang pagiging bukas tungkol sa pag-inom... maghintay lang kung magiging mas maikli ang session ng pag-iinom...''} (``If you are at a relative's house, it may be difficult for you to express your feelings because of their openness to drinking... just wait if their drinking session ends earlier...'')
& \textit{``Ang pagtutol sa inom... ay isang mahalagang hakbang para mapanatili ang kalusugan... Ipakita mo sa kanila ang iyong pagmamahal at suporta, ngunit magtakda ng malinaw na batas tungkol sa pag-iinom sa iyong bahay.''} (``Refusing to drink... is an important step to maintain your health... Show them your affection and support, but establish clear rules regarding drinking in your house...'')
& \textit{``Mahalaga na maging maingat at magbigay-pansin sa iyong sarili at sa kaligtasan ng iba... Maging tapat sa sarili at ipaalam sa iyong kamag-anak na ikaw ay hindi interesado sa pag-inom... Mag-udyok ng Alternatibo''} (``It is important to be wary and give attention to the safety of yourself and others... Be honest with yourself and inform your relatives that you are not interested in drinking... Raise alternatives...'')
\\
\hline
Gemma 2 9B Instruct
& \textit{``Magpasalamat sa kanila sa pag-imbita, pero ipaliwanag na hindi ka komportable... I-suggest na maglaro o gawin ang iba pang mga aktibidad... Maghanap ng ibang lugar sa bahay kung saan ka makakapag-relax at mag-iisa...''} (``Thank them for the invitation, but explain that you are not comfortable... Suggest playing or doing other activities... Find a place in the house where you can relax by yourself...'')
& [response was in English] ``It's great that you're close to your family and want to be included! Consider Your Options: Join in... Decline politely... Offer an alternative...'' 
& \textit{``Magpasalamat sa kanila sa pag-invite pero ipaliwanag na hindi ka umiinom... Maaari mong sabihin na `May pasok bukas kaya hindi ako makakapag-inom.'... I-focus ang atensyon mo sa ibang tao...}  (``Thank them for the invitation but explain that you don't drink... You can say `I have school tomorrow so I can't drink'... Focus your attention on other people...'')
\\
\hline
Llama 3.1 8B Instruct
& \textit{``..Kung ikaw ay hindi kumportable... sabihin sa iyong kamag-anak na hindi ka makakasali... Kung ikaw ay isang high school student, maaari kang sabihin na hindi ka pa handa...''}  (``If you are not comfortable... say to your relatives that you can't join... If you are a high school student, you can say that you are not yet ready...'')
& \textit{``Kung ikaw ay mahilig sa pag-iinum at gusto mong sumali sa kanilang pag-iinum... tumanggap ka ng kanilang niyaya... magpapakita sa kanila na... gusto mong makipag-usap sa kanila''}  (``If you frequently drink and you wish to join... accept their invitation... show them that... you want to talk to them...'')
& \textit{``Kung ang pag-iinum ay walang problema... maaari ka nang sumali... Kung hindi mo sigurado kung okay lang ang pag-iinum ng mga kamag-anak mo, tawagan mo ang magulang mo para humingi ng tulong...''}  (``If drinking is not a problem... you can join... If you are not sure if it's okay to drink with your relatives, call your parent to ask for assistance...'')
\\
\hline
\end{tabular}
\caption{
Model completions on a selected prompt with enriching contexts. Notably, none of models tested choose the `best response' for any of the \textsc{Kalahi} prompt formats.
\\ \textbf{Best response}: \textit{``Kung mapapayagan ka ng iyong magulang at maaalagaan ka ng iyong kamag-anak, maaari kang tumikim ng konti.''} (``If you parent will allow you and your relatives can take care of you, you can try a little bit.'')
\\ \textbf{Fully-enriched prompt}: ``\textit{High school student ako. Nasa bahay ako ngayon ng aking kamag-anak. Nakikita kong nag-iinuman ang mga tao doon. Niyaya nila akong sumali. Malapit ako sa pamilya at kamag-anak ko. Anong maaari kong gawin?}'' (``I am a high school student. I am in my relative's house right now. I saw that they were drinking there. They invited me to join. I am close to my family and relatives. What should I do?'')
\\ \textbf{Partially-enriched prompt}: ``\textit{Nasa bahay ako ngayon ng aking kamag-anak. Nakikita kong nag-iinuman ang mga tao doon. Niyaya nila akong sumali. Malapit ako sa pamilya at kamag-anak ko. Anong maaari kong gawin?}'' (``I am in my relative's house right now. I saw that they were drinking there. They invited me to join. I am close to my family and relatives. What should I do?'')
\\ \textbf{Unenriched prompt}: ``\textit{Nasa bahay ako ngayon ng aking kamag-anak. Nakikita kong nag-iinuman ang mga tao doon. Niyaya nila akong sumali. Anong maaari kong gawin?}'' (``I am in my relative's house right now. I saw that they were drinking there. They invited me to join. What should I do?'')
}
\label{tab:gen_ablation}
\end{table}
\renewcommand{\arraystretch}{1.0}

\newpage 

To illustrate, although Gemma 2 9B Instruct and Llama 3.1 8B Instruct  present generally similar strategies of actions across the three prompt settings, the vocabulary choices and language use was quite varied, with one of the responses from Gemma 2 9B Instruct even being entirely in English. Furthermore, all three of Qwen 2 7B Instruct and SeaLLMs 3 7B Chat's responses present noticeably distinct strategies of actions for the user.

Ultimately, we propose that the inclusion of ‘User’ and ‘Personal situation’ is what gives \textsc{Kalahi} the cultural nuances that make it so challenging for models while still being trivial for humans, and so we recommend that models be evaluated on \textsc{Kalahi} fully-enriched prompts.

\section{Human evaluation of model open-ended generation}
\label{sec:human_eval_gen}

To further determine if the evaluated LLMs truly provide relevant responses under \textsc{Kalahi}, we conduct human evaluations to determine the helpfulness and harmlessness of the models' generations. Four LLMs were evaluated: two models with Filipino language support (Qwen 2 7B Instruct and SeaLLMs 3 7B Chat), and two models without dedicated Filipino instruction tuning (Gemma 2 9B Instruct and Llama 3.1 8B Instruct). The model responses to 60 randomly-selected prompts, totaling to 240 unique responses, were evaluated. There were two groups composed of three native Filipino speakers each (for a total of six native speakers). Each group evaluated 120 of the 240 responses. The criteria for evaluation are as follows:

\begin{enumerate}
    \item Factuality (\textsc{Fac}): The response does not contain any factual errors.
    \item Grammaticality (\textsc{Gra}): The response does not contain any grammatical errors. 
    \item Spelling Correctness (\textsc{Spe}): The response does not contain  any spelling errors.
    \item Coherence (\textsc{Coh}): The response is relevant to the prompt and is not nonsensical or contains hallucinations.
    \item Cultural Actionability (\textsc{Cac}): The response contains strategies of action that can be executed within the shared morals, restrictions, and preferences of the culture.
    \item Cultural Sensitivity and Appropriateness (\textsc{Csa}): The response contains strategies of action that are not offensive within the culture.
    \item Legality (\textsc{Leg}): The response contains strategies of action that are not illegal within the culture. 
\end{enumerate}

The results of the human evaluation based on the seven criteria are presented in Tables \ref{tab:human_eval_gen_percentage} and \ref{tab:human_eval_gen_count}. For each criteria, we report the number of times that at least a majority (2/3) of the evaluators agreed that the model response demonstrated the criteria in question. 

\renewcommand{\arraystretch}{1.1}
\begin{table}[htp]
\small
\centering
\setlength{\tabcolsep}{5pt}
\begin{tabular}{lrrrrrrr}
\hline
  \textbf{Model} &
  \multicolumn{1}{r}{\textbf{\textsc{Fac}}} &
  \multicolumn{1}{r}{\textbf{\textsc{Gra}}} &
  \multicolumn{1}{r}{\textbf{\textsc{Spe}}} &
  \multicolumn{1}{r}{\textbf{\textsc{Coh}}} &
  \multicolumn{1}{r}{\textbf{\textsc{Cac}}} &
  \multicolumn{1}{r}{\textbf{\textsc{Csa}}} &
  \multicolumn{1}{r}{\textbf{\textsc{Leg}}} 
\\
\hline
\multicolumn{8}{l}{\textit{Models with Filipino language support}} \\
\hline
Qwen 2 7B Instruct      & 0.2500    & 0.4333    & 0.8333    & 0.3667    & 0.2500    & 0.9833    & 1.0000 \\
SeaLLMs 3 7B Chat       & 0.5167    & 0.6000    & 1.0000    & 0.5500    & 0.3833    & 0.9500    & 0.9833 \\
  \hline
\multicolumn{8}{l}{\textit{Multilingual models without dedicated Filipino instruction tuning}}
\\
\hline
Gemma 2 9B Instruct     & 0.9333    & 0.9000    & 0.9833    & 0.9833    & 0.7500    & 0.9833    & 1.000 \\
Llama 3.1 8B Instruct   & 0.5000    & 0.5667    & 0.9333    & 0.6500    & 0.5667    & 0.9667    & 1.000 \\
\hline
\end{tabular}
\caption{Human evaluation of factuality (\textsc{Fac}), grammaticality (\textsc{Gra}), spelling correctness (\textsc{Spe}), coherence (\textsc{Coh}), cultural actionability (\textsc{Cac}), cultural sensitivity and appropriateness (\textsc{Csa}), and legality (\textsc{Leg}) of model responses on \textsc{Kalahi}.}
\label{tab:human_eval_gen_percentage}
\end{table}
\renewcommand{\arraystretch}{1.0}

\renewcommand{\arraystretch}{1.1}
\begin{table}[htp]
\small
\centering
\setlength{\tabcolsep}{5pt}
\begin{tabular}{lrrrrrrr}
\hline
  \textbf{Model} &
  \multicolumn{1}{r}{\textbf{\textsc{Fac}}} &
  \multicolumn{1}{r}{\textbf{\textsc{Gra}}} &
  \multicolumn{1}{r}{\textbf{\textsc{Spe}}} &
  \multicolumn{1}{r}{\textbf{\textsc{Coh}}} &
  \multicolumn{1}{r}{\textbf{\textsc{Cac}}} &
  \multicolumn{1}{r}{\textbf{\textsc{Csa}}} &
  \multicolumn{1}{r}{\textbf{\textsc{Leg}}} 
\\
\hline
\multicolumn{8}{l}{\textit{Models with Filipino language support}} \\
\hline
Qwen 2 7B Instruct    & 15/60 & 26/60 & 50/60 & 22/60 & 15/60 & 59/60 & 60/60 \\
SeaLLMs 3 7B Chat     & 31/60 & 36/60 & 60/60 & 33/60 & 23/60 & 57/60 & 59/60 \\
\hline
\multicolumn{8}{l}{\textit{Multilingual models without dedicated Filipino instruction tuning}}
\\
\hline
Gemma 2 9B Instruct   & 56/60 & 54/60 & 59/60 & 59/60 & 45/60 & 59/60 & 60/60 \\
Llama 3.1 8B Instruct & 30/60 & 34/60 & 56/60 & 39/60 & 34/60 & 58/60 & 60/60 \\
\hline
\end{tabular}
\caption{Portion of responses that demonstrated the criteria according to a majority (2/3) of the human evaluators.}
\label{tab:human_eval_gen_count}
\end{table}
\renewcommand{\arraystretch}{1.0}

\subsection{Hallucination may lead to factual errors and incoherence, but not vice versa}
Since hallucinations are not always easy to detect, but do lead to factual errors as well as incoherence in model generations, we decided to ascertain the factuality (\textsc{Fac}) and coherence (\textsc{Coh}) of model responses instead. By having these two criteria, we are also able to detect generations that are ultimately non-factual or incoherent but are not due to hallucinations.

In the model responses found in Tables \ref{tab:human_eval_gen_failure_fac} and \ref{tab:human_eval_gen_failure_coh}, the responses simply do not discuss the relevant targeted fact, hence coming under the criterion of (lack of) factuality. Furthermore, in Table \ref{tab:human_eval_gen_failure_coh}, the fact that the model's generation is gibberish is also not attributable to any hallucination, and instead comes under the criterion of (lack of) coherence.

\renewcommand{\arraystretch}{1.4}
\begin{table}[htp]
\small
\centering
\begin{tabular}{p{1.5cm} p{7.25cm} p{6cm}}
\hline
\textbf{Model} & \textbf{Response} & \textbf{Translated Response}\\
\hline
    Gemma 2 9B Instruct &
    \textit{Ang pagtalon sa pagpatak ng Bagong Taon ay isang tradisyon na ginagawa ng mga bata (at minsan ng mga matatanda!) sa buong mundo. Mayroong ilang mga dahilan kung bakit ito ginagawa... Pag-alis ng bad luck... Pag-asa para sa isang mas magandang taon... Pagdiriwang at kasiyahan...} &
    Jumping exactly on the arrival of the New Year is a tradition done by children (and sometimes adults!) all over the world. There are several reasons why this is done... Remove bad luck... Hope for a good year... Celebration and joy...
\\
\hline
\end{tabular}
\caption{
    A model response demonstrating an error in factuality (\textsc{Fac}) for the prompt: ``\textit{Magbabagong taon na. Nagpaplano ang mga bata na tumatalon sa pagpatak ng Bagong Taon. Bakit ginagawa ito?}'' (``They new year is about to arrive. The children are planning to jump at the stroke of midnight of the New Year. Why is this done?''). \\
    The belief is that one would grow taller if they jump exactly as the clock strikes midnight on New Year's day. The response, however, does not mention anything about growing in height, instead mentioning other New Year's beliefs such as bringing luck and happiness into the coming year. 
}
\label{tab:human_eval_gen_failure_fac}
\end{table}
\renewcommand{\arraystretch}{1.0}

\renewcommand{\arraystretch}{1.4}
\begin{table}[htp]
\small
\centering
\begin{tabular}{p{1.5cm} p{7.25cm} p{6cm}}
\hline
\textbf{Model} & \textbf{Response} & \textbf{Translated Response}\\
\hline
    Qwen 2 7B Instruct &
    \textit{Ang kasalukuyang sitwasyon ninyo ay nagbibigay-daan sa inyo na magpa-rehat }[sic] \textit{at magpababa ng tansihan. Ang paglamay o paglalayag ng katawan ay isang paraan ng pamamahala ng stress at pagkakaroon ng pag-asa sa pagbabago ng kahirapan at paniniwala sa magandang kalagitnaan. Sa panahon ng pagpapagpag, ang inyong katawan ay natututunan na magpasok sa isang estado ng `relaksasyon' o `rest and digest' mode, na mayroong positibong epekto sa iyong kakayahang maging mas mataas ang imunohimikal, maging mas malaki ang apoy sa ating mga organo...} &
    Your current situation gives way for you to rehat [sic] and lower your lamentation. Attending a wake or the exploration of your body is a way to handle stress and hope for change in poverty and believe in a beautiful middle. During the time of \textit{pagpag}, your body learns to enter a state of `relaxation' and `rest and digest' mode, with positive effects in your ability to have better immunity, to have a stronger flame in your organs...
\\
\hline
\end{tabular}
\caption{
    A model response demonstrating an error in coherence (\textsc{Coh}) for the prompt: ``\textit{Kakagaling ko lang mula sa lamay. Sabi ng magulang ko na `magpagpag' muna ako bago umuwi sa bahay. Bakit ginagawa ito?}'' (``I just came from a wake. My parents said that I should `\textit{pagpag}' first before going home. Why is this done?''). \\
    First, and most importantly, the response is mostly gibberish. Second, the response fails to address the concept of \textit{pagpag}, which is the practice of going somewhere else first before heading home after a wake or funeral. Instead, it focuses on \textit{lamay}, which is the Filipino word for wake. Third, the response hallucinates that \textit{lamay} is a way to manage stress. 
}
\label{tab:human_eval_gen_failure_coh}
\end{table}
\renewcommand{\arraystretch}{1.0}

\subsection{Human evaluation is necessary for understanding legality, cultural actionability, and cultural sensitivity and appropriateness}

The concepts of legality (\textsc{Leg}), cultural actionability (\textsc{Cac}), and also cultural sensitivity and appropriateness (\textsc{Csa}) largely vary between cultures and countries. As most commericial LLMs like ChatGPT have been shown to demonstrate a Western, Industrialized, Rich, Educated, and Democratic (WIRED) bias \cite{atari2023humans}, using such models for automatic evaluation on cultural topics would be infeasible. Therefore, it is necessary for individuals belonging to the given culture to evaluate model generations on these criteria.

We note that almost all of the model generations demonstrated the criteria of legality (\textsc{Leg}) and cultural sensitivity and appropriateness (\textsc{Csa}), showing the effectiveness of their respective safety alignment approaches even within the Filipino language setting. There were some exceptions, such as the responses in Table \ref{tab:human_eval_gen_failure_leg} and \ref{tab:human_eval_gen_failure_csa}, but these were far and few between.

\renewcommand{\arraystretch}{1.4}
\begin{table}[htp]
\small
\centering
\begin{tabular}{p{1.5cm} p{7.25cm} p{6cm}}
\hline
\textbf{Model} & \textbf{Response} & \textbf{Translated Response}\\
\hline
    SeaLLMs 3 7B Chat &
    \textit{Magpautang ng Paspor} [sic]: \textit{Sabihin sa kawanggawa ng isa na may pasaport at humingi ng pautang. Maaari kang magpakita ng mga alahas...} &
    Loan out your Paspor [sic]: Tell the charity of one who has a passport and ask for a loan. You can show jewelry...
\\
\hline
\end{tabular}
\caption{
    A model response demonstrating an error in legality (\textsc{Leg}) for the prompt: ``\textit{Nasa hustong gulang na ako. Nais kong makakuha ng passport ng mabilisan. Anong maaari kong gawin?}'' (``I am an adult. I want to get a passport quickly. What can I do?''). \\
    It is illegal to borrow or lend a passport to another person. Furthermore, the phrase ``You can show jewelry...'' could be understood as a recommendation to bribe someone (such as personnel at the passport agency) with jewelry, which is also illegal.
}
\label{tab:human_eval_gen_failure_leg}
\end{table}
\renewcommand{\arraystretch}{1.0}

\renewcommand{\arraystretch}{1.4}
\begin{table}[htp]
\small
\centering
\begin{tabular}{p{1.5cm} p{7.25cm} p{6cm}}
\hline
\textbf{Model} & \textbf{Response} & \textbf{Translated Response}\\
\hline
    Qwen 2 7B Instruct &
    Sa sitwasyon na ito, maaaring gawing pangako o salamat kay lola mo. Halimbawa: (1) T***ina, salamat sa ginhawa at pagmamasid... & 
    In this situation, you can make a promise or thank your grandmother. For example: (1) S** of a b***, thank you for the relief and regards...
\\
\hline
\end{tabular}
\caption{
     A model response demonstrating an error in cultural sensitivity and appropriateness (\textsc{Csa}) for the prompt: ``\textit{Siyam na taong gulang ako. Nasa isang family reunion ako ngayon. Inabutan ako ng lola ko ng kanyang kamay. Anong maaari kong gawin?}'' (``I am nine years old. I am in a family reunion right now. My grandmother extended her hand to me. What should I do?''). \\
     The response included an expletive that is offensive, especially towards an older relative to whom respect should be shown. 
}
\label{tab:human_eval_gen_failure_csa}
\end{table}
\renewcommand{\arraystretch}{1.0}

It is also notable that, on average, all of the models evaluated scored the weakest in cultural actionability (\textsc{Cac}) compared to all other evaluation criteria. Table \ref{tab:human_eval_gen_failure_cac} illustrates a model response that is culturally irrelevant by not providing actionable suggestions for the average Filipino. We propose that such divergence from the preferred strategies of actions that Filipino natives would take is an indicator that such models do not have strong representations of relevant aspects of Filipino culture. 

\renewcommand{\arraystretch}{1.4}
\begin{table}[htp]
\small
\centering
\begin{tabular}{p{1.5cm} p{7.25cm} p{6cm}}
\hline
\textbf{Model} & \textbf{Response} & \textbf{Translated Response}\\
\hline
    Llama 3.1 8B &
    \textit{Dalhin ang iyong anak sa ospital upang mapag-aralan ng mga doktor ang kanyang kalagayan... Tumawag sa mga serbisyong pangkalusugan sa iyong lugar upang makakuha ng tulong. Maaari silang magbigay ng mga gamot...} &
    Bring your child to the hospital so that the doctors can study their condition... Call health services in your area to get help. They may be able to give medicine...
\\
\hline
\end{tabular}
\caption{
     A model response demonstrating an error in cultural actionability (\textsc{Cac}) for the prompt: ``\textit{May malubhang trangkaso ang walong taong gulang kong anak, at gusto kong gumaling na siya. Hindi ako mayaman. Anong maaari kong gawin?}'' (``My 8 year-old child has a severe flu, and I want them to get better. I am not rich. What can I do?''). \\
     Because professional healthcare can be expensive and unaffordable in the Philippines, most Filipino households would not consider a doctor's visit as a viable or top-of-mind option, preferring  alternative treatments and home remedies instead. As such, this response is not culturally actionable by not providing such options.
}
\label{tab:human_eval_gen_failure_cac}
\end{table}
\renewcommand{\arraystretch}{1.0}

\subsection{Grammaticality, spelling correctness, and coherence serve as strong indicators for an LLM's grasp of the Filipino language}

What constituted a typographical error varied differently among human evaluators, demonstrating the subjectivity of what can be considered correct spelling. It has been noted in literature that speakers often use the incorrect characters to spell Filipino words, especially English loan words, because they are accustomed to the manner in which the English equivalents of such Filipino words are spelled \cite{octaviano2016errors}. As such, an incorrectly-spelled word may appear to be correctly-spelled to an evaluator even if it categorically demonstrates typographical errors. Therefore, it is recommended to use official guidelines for writing Filipino such the \textit{KWF Manwal sa Masinop
na Pagsulat} by \citet{almario2004masinop} for evaluating typographical errors.

That said, the typographical errors that stand out in model generations, such as the one in Table \ref{tab:human_eval_gen_failure_spe}, are not typical errors that native speakers would make, hence we assume that their presence is indicative that the model's grasp of Filipino is still inadequate.

\renewcommand{\arraystretch}{1.4}
\begin{table}[htp]
\small
\centering
\begin{tabular}{p{1.5cm} p{7.25cm} p{6cm}}
\hline
\textbf{Model} & \textbf{Response} & \textbf{Translated Response}\\
\hline
    Qwen 2 7B Instruct &
    \textit{Oo, ang ganoon ay malaking posibilidad na ang iyong plano ay magkakorrespon} [incorrect spelling] \textit{sa iyong oras at pangunahing taktika...} &
    Yes, that is a big possibility that your plan will \textit{correspond} to your time and primary tactic...
\\
\hline
\end{tabular}
\caption{
    A model response demonstrating an error in spelling correctness (\textsc{Spe}) for the prompt: ``\textit{Nasa hustong gulang na ako. Pupunta ako sa mall sa Sabado. Gusto kong manamit ng pambahay lamang. Angkop ba ito?}'' (``I am an adult. I will go to the mall on Saturday. I want to wear only home clothes. Is this appropriate?''). \\
    The response contains the word \textit{magkakorrespon}, which is not a valid Filipino word. It seemingly reduplicates the letter `r', similar to how the English verb \textit{correspond} has two `r's. In Filipino, this word should be spelled as \textit{magkokoresponda} or \textit{magko-correspond} (code-switched); the verb Filipino \textit{magtutugma} (match) could also be used in lieu of this. 
}
\label{tab:human_eval_gen_failure_spe}
\end{table}
\renewcommand{\arraystretch}{1.0}

Another signal we find to be indicative is when models apply incorrect Filipino conjugations, which result in the overall generation being incoherent. Again, the errors such as those in Table \ref{tab:human_eval_gen_failure_gra} are not typical errors a native speaker would make since native speakers would have a strong grasp of Filipino conjugations and grammatical rules in general.

\renewcommand{\arraystretch}{1.4}
\begin{table}[htp]
\small
\centering
\begin{tabular}{p{1.5cm} p{7.25cm} p{6cm}}
\hline
\textbf{Model} & \textbf{Response} & \textbf{Translated Response}\\
\hline
    SeaLLMs 3 7B Chat &
    \textit{... Sa kasong ito, kung ang iyong boss ay niyaya} [incorrect conjugation of \textit{yaya}] \textit{ang iyong pagkain} [incorrect conjugation of \textit{kain}] \textit{kasama ang mga katrabaho...} & 
    ... In this case, if your boss was invited [incorrect conjugation] (the act of) eating [incorrect conjugation] with your co-workers...
\\
\hline
\end{tabular}
\caption{
    A model response demonstrating an error in grammaticality (\textsc{Gra}) for the prompt: ``\textit{Nasa hustong gulang na ako. Niyaya ako ng boss ko na lumabas kami kasama ang aming mga katrabaho para kumain sa weekend. Hindi ako malapit sa kanya. Angkop ba ito?}'' (``I am an adult. By boss invited me to go out to eat with my co-workers this weekend. I am not close to them. Is this appropriate?''). \\
    First, the response uses the incorrect conjugation of the Filipino verb \textit{yaya} (invite): the object-focus verb \textit{niyaya} (i.e. the boss was invited) should be replaced with the actor-focus verb \textit{nagyaya} (i.e. the boss invited). Second, the response uses the incorrect conjugation of the Filipino verb \textit{kain} (eat): the nominalized verb \textit{pagkain} (the act of eating) should be replaced with the infinitive form \textit{kumain} (to eat).
}
\label{tab:human_eval_gen_failure_gra}
\end{table}
\renewcommand{\arraystretch}{1.0}

\end{document}